\documentclass{article}

\usepackage[numbers]{natbib}
\usepackage[preprint]{neurips_2024}

\usepackage[dvipsnames]{xcolor}         
\definecolor{linkColor}{rgb}{0.18,0.39,0.62}
\usepackage[utf8]{inputenc} 
\usepackage[T1]{fontenc}    
\usepackage[colorlinks=true,linkcolor=linkColor,citecolor=linkColor,filecolor=linkColor,urlcolor=linkColor]{hyperref}       
\usepackage{url}            
\usepackage{booktabs}       
\usepackage{amsfonts}       
\usepackage{nicefrac}       
\usepackage{microtype}      

\usepackage{graphicx}
\usepackage{arydshln}

\usepackage{booktabs}
\usepackage{multirow}
\usepackage{caption}
\usepackage{subcaption}
\usepackage{makecell}
\usepackage{csquotes}
\usepackage{epigraph}

\RequirePackage{algorithm}
\RequirePackage{algorithmic}

\usepackage{multirow}
\usepackage{amsmath}
\usepackage{capt-of}
\usepackage{tabularx}
\usepackage{epsfig}
\usepackage{amssymb}
\usepackage{amsfonts}
\usepackage{booktabs}
\usepackage{scalerel}
\usepackage[inline]{enumitem}
\usepackage{listings}
\usepackage{varwidth}
\usepackage[export]{adjustbox}
\usepackage{tikz}
\usetikzlibrary{tikzmark}

\usepackage{stmaryrd}
\usepackage{bbm}
\usepackage{wrapfig}
\usepackage{pifont}
\usepackage[noabbrev]{cleveref}

\newcommand{\mypara}[1]{\textbf{#1}~~}

\definecolor{deepblue}{rgb}{0,0,0.5}
\definecolor{officeblue}{RGB}{0,102,204}
\definecolor{deepred}{rgb}{0.6,0,0}
\definecolor{deepgreen}{rgb}{0,0.5,0}
\definecolor{mybrickred}{RGB}{182,50,28}

\definecolor{fillcolor}{RGB}{216,217,252}

\usepackage{etoolbox}
\usepackage{framed}

\newif\ifxetexorluatex
\ifxetex
  \xetexorluatextrue
\else
  \ifluatex
    \xetexorluatextrue
  \else
    \xetexorluatexfalse
  \fi
\fi
\newcommand*\quotesize{60} 
\newcommand*{\openquote}
   {\tikz[remember picture,overlay,xshift=-4ex,yshift=-2.5ex]
   \node (OQ) {\fontsize{\quotesize}{\quotesize}\selectfont``};\kern0pt}

\newcommand*{\closequote}[1]
  {\tikz[remember picture,overlay,xshift=4ex,yshift={#1}]
   \node (CQ) {\fontsize{\quotesize}{\quotesize}\selectfont''};}

\colorlet{shadecolor}{white}

\newcommand*\shadedauthorformat{\emph} 

\newcommand*\authoralign[1]{%
  \if#1l
    \def\authorfill{}\def\quotefill{\hfill}
  \else
    \if#1r
      \def\authorfill{\hfill}\def\quotefill{}
    \else
      \if#1c
        \gdef\authorfill{\hfill}\def\quotefill{\hfill}
      \else\typeout{Invalid option}
      \fi
    \fi
  \fi}
{\authoralign{#1}
\ifblank{#2}
   {\def\shadequoteauthor{}\def\yshift{-2ex}\def\quotefill{\hfill}}
   {\def\shadequoteauthor{\par\authorfill\shadedauthorformat{#2}}\def\yshift{2ex}}
\begin{snugshade}\begin{quote}\openquote}
{\shadequoteauthor\quotefill\closequote{\yshift}\end{quote}\end{snugshade}}

\newfloat{Code}{htbp}{Code}

\usepackage{amsmath,amsfonts,bm}

\def\eqref#1{equation~(\ref{#1})}

\def\1{\bm{1}}

\def\vx{{\bm{x}}}

\DeclareMathAlphabet{\mathsfit}{\encodingdefault}{\sfdefault}{m}{sl}
\SetMathAlphabet{\mathsfit}{bold}{\encodingdefault}{\sfdefault}{bx}{n}

\newcommand{\softmax}{\mathrm{softmax}}

\newcommand\our{{YOCO}}

\newcommand\gretnet{gRetNet}

\title{You Only Cache Once:\\Decoder-Decoder Architectures for Language Models}

\author{
Yutao Sun\thanks{~Equal contribution. $\diamond$ Corresponding author.}$~~^{\dag\ddag}$~~~~Li Dong\footnotemark[1]$~~^{\dag}$~~~~Yi Zhu$^{\dag}$~~~Shaohan Huang$^{\dag}$ \\
\bf Wenhui Wang$^{\dag}$~~~Shuming Ma$^{\dag}$~~~\bf Quanlu Zhang$^{\dag}$~~~Jianyong Wang$^{\ddag}$~~~Furu Wei$^{\dag}$$^{\diamond}$ \\
$^\dag$ Microsoft Research ~~~~
$^\ddag$ Tsinghua University \\
{\href{https://aka.ms/GeneralAI}{https://aka.ms/GeneralAI}}
}

\begin{document}

\maketitle

\begin{abstract}
We introduce a decoder-decoder architecture, \our{}, for large language models, which only caches key-value pairs once. It consists of two components, i.e., a \textit{cross-decoder} stacked upon a \textit{self-decoder}. The self-decoder efficiently encodes global key-value (KV) caches that are reused by the cross-decoder via cross-attention. The overall model behaves like a decoder-only Transformer, although \our{} only caches once. The design substantially reduces GPU memory demands, yet retains global attention capability. Additionally, the computation flow enables prefilling to early exit without changing the final output, thereby significantly speeding up the prefill stage. Experimental results demonstrate that \our{} achieves favorable performance compared to Transformer in various settings of scaling up model size and number of training tokens. We also extend \our{} to 1M context length with near-perfect needle retrieval accuracy. The profiling results show that \our{} improves inference memory, prefill latency, and throughput by orders of magnitude across context lengths and model sizes. Code is available at \url{https://aka.ms/YOCO}.
\end{abstract}

\vfill{}

\begin{figure*}[ht]
\centering
\captionsetup{type=figure}
\includegraphics[width=0.92\textwidth]{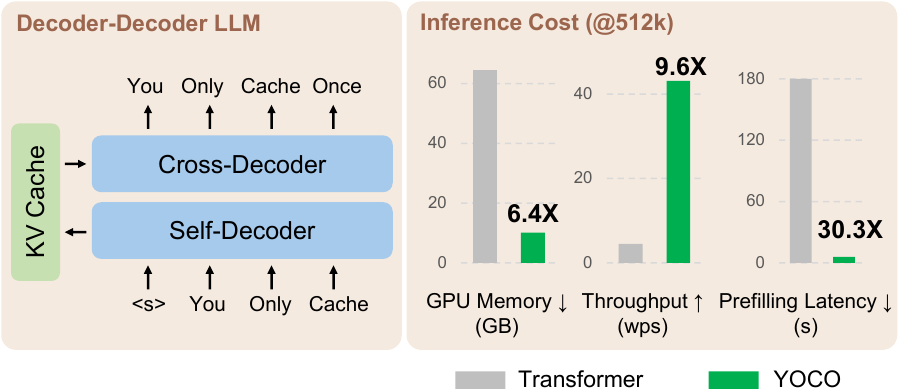}
\caption{We propose a decoder-decoder architecture, \our{}, for large language model, which only caches key/value once.
\our{} markedly reduces the KV cache memory and the prefilling time, while being scalable in terms of training tokens, model size, and context length.
The inference cost is reported to be 512K as the context length, and \Cref{fig:memory,fig:cache:mem:per:token,fig:prefilling,fig:throughput} present more results for different lengths.
}
\label{fig:highlight}
\end{figure*}

\vfill{}

\newpage
\section{Introduction}
\label{sec:intro}

The decoder-only Transformer~\cite{transformer} has become the de facto architecture for language models.
Numerous efforts have continued to develop suitable architectures for language modeling.
There have been main strands of explorations.
First, encoder-only language models, such as BERT~\cite{bert}, bidirectionally encode the input sequence.
Second, encoder-decoder models, such as T5~\cite{t5}, use a bidirectional encoder to encode input and a unidirectional decoder to generate output.
Both of the above layouts struggle with autoregressive generation due to bidirectionality.
Specifically, encoders have to encode the whole input and output tokens again for the next generation step.
Although encoder-decoder can use only decoder to generate, the output tokens do not fully leverage the parameters of encoder, especially for multi-turn conversation.
Third, decoder-only language models, such as GPT~\cite{gpt3}, generate tokens autoregressively.
By caching the previously computed key/value vectors, the model can reuse them for the current generation step.
The key-value (KV) cache avoids encoding the history again for each token, greatly improving the inference speed.
This compelling feature establishes the decoder-only language model as the standard option.

However, as the number of serving tokens increases, the KV caches occupy a lot of GPU memory, rendering the inference of large language models memory-bounded~\cite{scaling:inference}.
For the example of a 65B-size language model (augmented with grouped-query attention~\cite{gqa} and 8-bit KV quantization), 512K tokens occupy about 86GB GPU memory, which is even larger than the capacity of one H100-80GB GPU.
In addition, the prefilling latency of long-sequence input is extremely high.
For instance, using four H100 GPUs, the 7B language model (augmented with Flash-Decoding~\cite{flashdec} and kernel fusion) requires about 110 seconds to prefill 450K tokens, and 380 seconds for 1M length.
The above bottlenecks make it difficult to deploy long-context language models in practice.

In this work, we propose a decoder-decoder architecture, \our{}, for large language models, which only caches KV pairs once.
Specifically, we stack cross-decoder upon self-decoder.
Given an input sequence, the self-decoder utilizes efficient self-attention to obtain KV caches.
Then the cross-decoder layers employ cross-attention to reuse the shared KV caches.
The decoder-decoder architecture is conceptually similar to encoder-decoder, but the whole model behaves more like a decoder-only model from the external view.
So, it naturally fits into autoregressive generation tasks, such as language modeling.
First, because \our{} only caches once\footnote{The word ``once'' refers to global KV cache. Strictly, self-decoder also needs to store a certain number of caches. As the self-decoder utilizes an efficient attention module, the cache size is bounded to a constant, which can be ignored compared to global caches when the sequence length is large.}, the GPU memory consumption of KV caches is significantly reduced.
Second, the computation flow of the decoder-decoder architecture enables prefilling to early exit before entering the self-decoder.
The nice property speeds up the prefill stage dramatically, improving user experience for long-context language models.
Third, \our{} allows for more efficient system design for distributed long-sequence training.
In addition, we propose gated retention for self-decoder, which augments retention~\cite{retnet} with a data-controlled gating mechanism.

We conduct extensive experiments to show that \our{} achieves favorable language modeling performance and has many advantages in terms of inference efficiency.
Experimental results demonstrate that \our{} can be scaled up with more training tokens, larger model size, and longer context length.
Specifically, we scale up the 3B \our{} model to trillions of training tokens, attaining results on par with prominent Transformer language models, such as StableLM~\cite{stablelm}.
Moreover, the scaling curves ranging from 160M to 13B show that \our{} are competitive compared to Transformer.
We also extend the context length of \our{} to 1M tokens, achieving near perfect needle retrieval accuracy.
In the multi-needle test, \our{} obtains competitive results even compared to larger Transformers.

In addition to good performance on various tasks, the profiling results show that \our{} improves the GPU memory footprint, prefill latency, throughput, and serving capacity.
In particular, the memory of KV caches can be reduced by about $80\times$ for 65B models.
Even for a 3B model, the overall inference memory consumption can be reduced by two times for 32K tokens and by more than nine times for 1M tokens.
The prefill stage is speeded up by $71.8\times$ for the 1M context and $2.87\times$ for the 32K input.
For example, for a 512K context, \our{} reduces the Transformer prefilling latency from 180 seconds to less than six seconds.
The results position \our{} as a strong candidate model architecture for future large language models with native long-sequence support.

\section{You Only Cache Once (YOCO)}
\label{sec:arch}

\begin{figure*}[t]
\centering
\captionsetup{type=figure}
\includegraphics[width=0.62\textwidth]{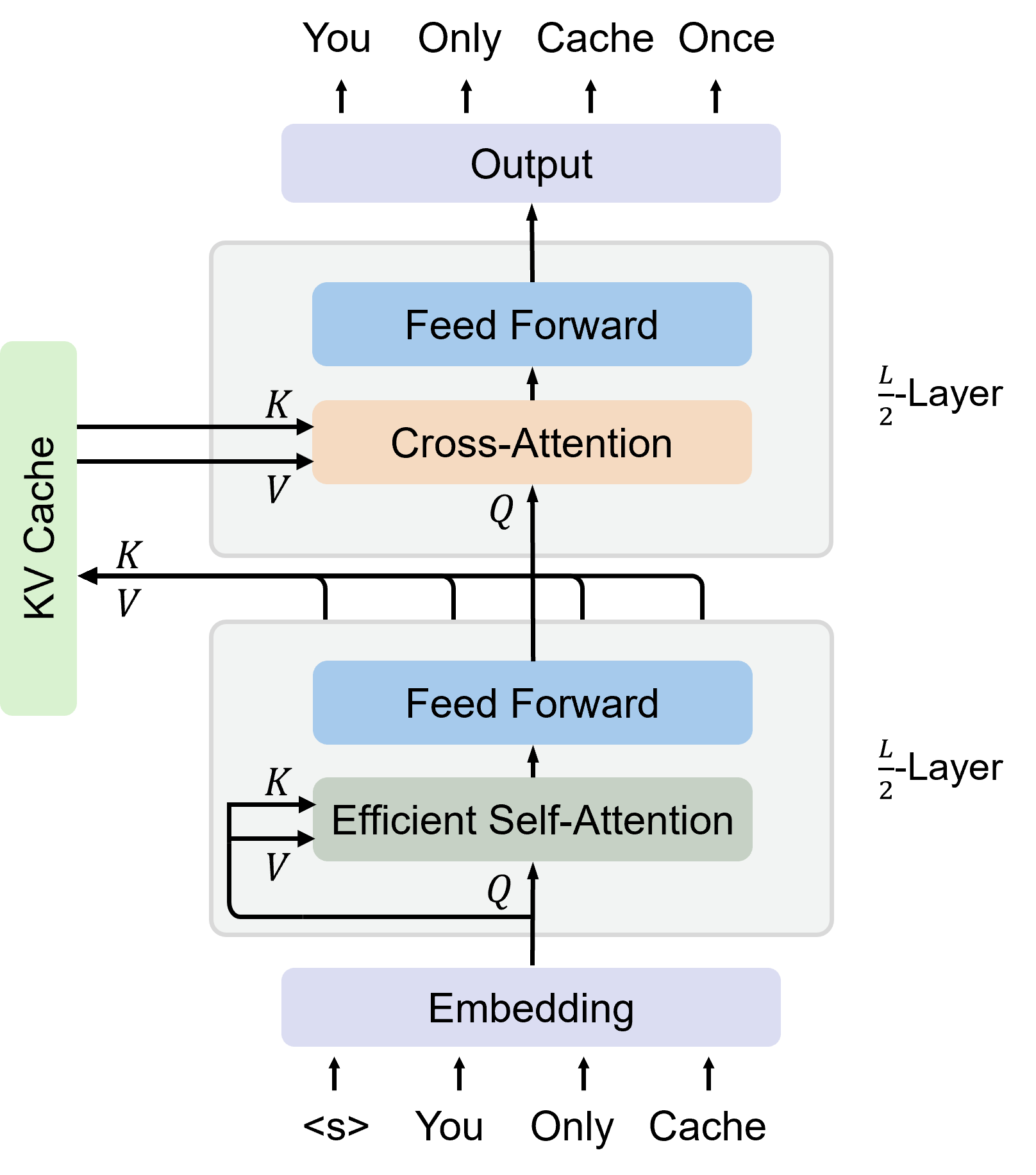}
\caption{Overview of the decoder-decoder architecture. Self-decoder generates the global KV cache. Then cross-decoder employs cross-attention to reuse the shared KV caches.
Both self-decoder and cross-decoder use causal masking.
The overall architecture behaves like a decoder-only Transformer, autoregressively generating tokens.
}
\label{fig:yoco}
\end{figure*}

The proposed architecture, named \our{}, is designed for autoregressive modeling, such as large language models (LLMs).
As shown in \Cref{fig:yoco}, the decoder-decoder architecture has two parts, i.e., self-decoder and cross-decoder.
Specifically, \our{} is stacked with $L$ blocks, where the first $\frac{L}{2}$ layers are self-decoder while the rest modules are cross-decoder.
%
Given an input sequence $x = x_1 \cdots x_{|x|}$, the input embeddings are packed into $X^0 = [\vx_1, \cdots, \vx_{|x|}] \in \mathbb{R}^{|x|\times d_\text{model}}$, where $d_\text{model}$ is hidden dimension.
We first obtain contextualized vector representations $X^{l} = \operatorname{Self-Decoder}(X^{l-1}), l \in [1, \frac{L}{2}]$, where $X^{\nicefrac{L}{2}}$ is used to produce KV caches $\hat{K}, \hat{V}$ for cross-decoder.
Then we compute $X^{l} = \operatorname{Cross-Decoder}(X^{l-1}, \hat{K}, \hat{V}), l \in [\frac{L}{2}+1, L]$ to get the output vectors $X^{L}$.

Both self- and cross-decoder follow a similar block layout (i.e., interleaved attention and feed-forward network) as in Transformer~\cite{transformer}. We also include pre-RMSNorm~\cite{rmsnorm}, SwiGLU~\cite{glu}, and grouped-query attention~\cite{gqa} as improvements.
The difference between the two parts lies in attention modules.
Self-decoder (\Cref{sec:self:decoder}) uses efficient self-attention (e.g., sliding-window attention).
In comparison, cross-decoder (\Cref{sec:cross:decoder}) uses global cross-attention to attend to the shared KV caches produced by the output of the self-decoder.

\subsection{Self-Decoder}
\label{sec:self:decoder}

Self-decoder takes token embeddings $X^0$ as input and compute intermediate vector representation $M=X^{\nicefrac{L}{2}}$:
\begin{equation}
\begin{aligned}
Y^{l}&=\operatorname{ESA}(\operatorname{LN}(X^{l}))+X^{l}\\
X^{l+1}&=\operatorname{SwiGLU}(\operatorname{LN}(Y^{l}))+Y^{l}\\
\end{aligned}
\end{equation}
where $\operatorname{ESA}(\cdot)$ represents efficient self-attention, $\operatorname{SwiGLU}(X) = (\operatorname{swish}(X W_G)\odot X W_1)W_2$, and RMSNorm~\cite{rmsnorm} is used for $\operatorname{LN}(\cdot)$.
Causal masking is used for efficient self-attention.

The key property of the efficient self-attention module is $\mathcal{O}(1)$ inference memory, i.e., constant number of KV caches.
For example, the cache size of sliding-window attention~\cite{sparsetransformer} depends on the window size instead of the input length.
More design choices (e.g., gated retention) of the efficient self-attention module are detailed in \Cref{sec:design:self:decoder}.

\subsection{Cross-Decoder}
\label{sec:cross:decoder}

First, the output of the self-decoder $X^{\nicefrac{L}{2}}$ generates global KV caches $\hat{K}, \hat{V}$ for cross-decoder:
\begin{equation}
\hat{K} = \operatorname{LN}(X^{\nicefrac{L}{2}}) W_K ,\quad \hat{V} = \operatorname{LN}(X^{\nicefrac{L}{2}}) W_V
\end{equation}
where $W_K, W_V \in \mathbb{R}^{d\times d}$ are learnable weights.
Then, cross-decoder layers are stacked after the self-decoder to obtain the final output vectors $X^{L}$.
The KV caches $\hat{K}, \hat{V}$ are reused by all the $\frac{L}{2}$ cross-decoder modules:
\begin{equation}
\begin{aligned}
\hat{Q}^{l} &= \operatorname{LN}(X^{l}) W_Q^l \\
Y^{l}&=\operatorname{Attention}(\hat{Q}^{l}, \hat{K}, \hat{V})+X^{l}\\
X^{l+1}&=\operatorname{SwiGLU}(\operatorname{LN}(Y^{l}))+Y^{l}\\
\label{eq:cross:decoder}
\end{aligned}
\end{equation}
where $\operatorname{Attention}(\cdot)$ is standard multi-head attention~\cite{transformer}, and $W_Q^l \in \mathbb{R}^{d\times d}$ is a learnable matrix.
Causal masking is also used for cross-attention.
Because cross-attention is compatible with group query attention~\cite{gqa}, we can further save the memory consumption of KV caches.
After obtaining $X^L$, a $\softmax$ classifier performs next-token prediction.

\subsection{Inference Advantages}
\label{sec:infer:adv}

In addition to competitive language modeling results, \our{} significantly reduces serving costs and improves inference performance.
We report detailed inference comparisons in \Cref{sec:exp:infer}.

\begin{table*}[t]  
\centering  
\begin{minipage}[b]{0.52\linewidth}  
\centering  
\captionsetup{type=figure}
\includegraphics[width=\textwidth]{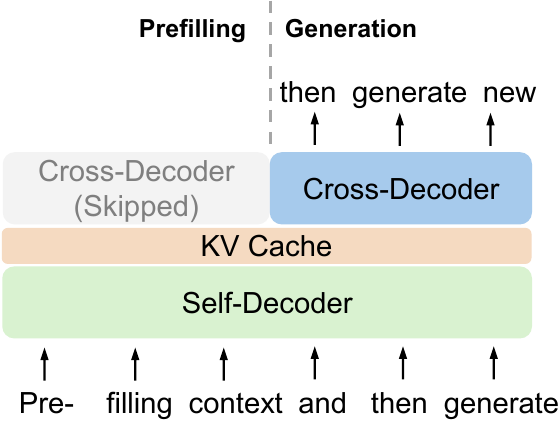}
\caption{\our{} Inference.
\textbf{Prefill}: encode input tokens in parallel.
\textbf{Generation}: decode output tokens one by one.
The computation flow enables prefilling to early exit without changing the final output, thereby significantly speeding up the prefill stage.
}
\label{fig:arch:inference}
\end{minipage}
\hfill  
\begin{minipage}[b]{0.47\linewidth}
\centering
\begin{tabular}{lc}
\toprule
& \textbf{KV Cache Memory}  \\
\midrule
Transformer & $\mathcal{O}(LND)$ \\
\our{} & $\mathcal{O}((N+L)D)$ \\
\bottomrule
\end{tabular}
\caption{Inference memory complexity of KV caches. $N, L, D$ are the sequence length, number of layers, and hidden dimension.}
\label{tbl:complexity:memory}
\begin{tabular}{lc}
\toprule
& \textbf{Prefilling Time} \\
\midrule
Transformer & $\mathcal{O}(LN^2D)$ \\
\our{} & $\mathcal{O}(LND)$ \\
\bottomrule
\end{tabular}
\caption{Prefilling time complexity of attention modules. $N, L, D$ are the same as above.}  
\label{tbl:complexity:prefill}
\end{minipage}
\end{table*}  

\mypara{Saving GPU Memory and Serving More Tokens.}
\Cref{tbl:complexity:memory} compares the memory complexity between Transformers and \our{}.
Specifically, because global KV caches are reused and efficient self-attention needs constant caches, the number of caches is $\mathcal{O}(N + CL)$, where $N$ is the input length, $C$ is a constant (e.g., sliding window size), and $L$ is the number of layers.
For long sequences, $CL$ is much smaller than $N$, so about $\mathcal{O}(N)$ caches are required, i.e., you only cache once.

In comparison, Transformer decoders have to store $N \times L$ keys and values during inference.
So \our{} roughly saves $L$ times GPU memory for caches compared to Transformer decoders.
Because the inference capacity bottleneck becomes KV caches (\Cref{fig:memorybar}), our method enables us to serve many more tokens without being out of GPU memory.
The increased batch size is also beneficial to inference throughput.

\mypara{Reducing Prefilling Time and Improving Throughput.}
As shown in \Cref{fig:arch:inference}, because the cross-decoder reuses the outputs of self-decoder, we can exit early before entering the cross-decoder during the prefill stage.
The intriguing property of computation dependency greatly accelerates the prefilling speed.

First, only half the layers are needed for forward computation, i.e., at least half prefilling latency reduction.
Second, the efficient attention modules of the self-decoder are usually fast.
For the example of 512K context length, we can decrease the prefilling latency from 180 seconds (Transformer with optimized inference, such as Flash-Decoding and kernel fusion) to less than 6 seconds (\Cref{fig:prefilling}).
Even for 32K length, \our{} has about three times speedup in terms of prefilling time.
\Cref{tbl:complexity:prefill} compares prefilling time complexity of attention modules between Transformer and \our{}.

\section{Design Choices of Self-Decoder}
\label{sec:design:self:decoder}

We can choose various efficient self-attention methods for self-decoder.
As long as the module only requires constant inference memory, the cache memory complexity of the self-decoder depends on the number of layers.
Moreover, a good module choice improves both training and deployment costs.
In this work, we use gated retention (\Cref{sec:gret}) or sliding-window attention (\Cref{sec:swa}).

\subsection{Gated Retention}
\label{sec:gret}

Gated retention (gRet, aka gRetNet or RetNet-3) augments retention~\cite{retnet} with a data-dependent gating mechanism, which achieves training parallelism, good performance, and low inference cost simultaneously for sequence modeling.
We use gRet as the default efficient self-attention module in the experiments.
The method unifies the parallel, recurrent, and chunkwise recurrent computation paradigms. These three representations are equivalent and can obtain the same computation results.
The training process usually uses the parallel or chunkwise recurrent paradigms, while the inference stage can employ the recurrent paradigm for constant KV memory.
We describe the three representations as follows:

\mypara{The Parallel Representation}
The gated retention is defined as:
\begin{equation}
\begin{aligned}
Q = (X W_Q) \odot \Theta ,& \quad K = (X W_K) \odot \overline{\Theta} ,\quad V = X W_V ,\quad \Theta_n = e^{in\theta} \\
\gamma = \operatorname{sigmoid} &(X W_{\gamma})^{1/\tau} , \quad D_{nm} =
\left\{
\begin{aligned}
& \prod_{i=m+1}^{n}\gamma_i, &n\ge m \\
& 0, &n < m \\
\end{aligned}
\right. \\
&\operatorname{gRet}(X) = (Q K^\intercal \odot D)V
\label{eq:gret:parallel}
\end{aligned}
\end{equation}
where $W_Q, W_K, W_V \in \mathbb{R}^{d\times d}$ and $W_{\gamma} \in \mathbb{R}^{d\times 1}$ are learnable weights, and the temperature term $\tau$ encourages $\gamma$ to 1 for better memorization~\cite{gla}.
The data-controlled decay is head-wise~\cite{gateloop} rather than element-wise so that the computation can fully utilize NVIDIA tensor cores.
Refer to \cite{retnet} for more details about the other designs.

\mypara{The Recurrent Representation}
Being equivalent to \Cref{eq:gret:parallel}, the output of gated retention can be computed recurrently.
For the $n$-th timestep, the output is obtained via:
\begin{equation}
\begin{aligned}
&S_n = \gamma_n S_{n-1} + K_n^{\intercal} V_n \\
&\operatorname{gRet}(X_n) = Q_n S_n, \quad n = 1, \cdots, |x| \\
\label{eq:gret:recurrent}
\end{aligned}
\end{equation}
where $Q, K, V, \gamma$ are the same as in \Cref{eq:gret:parallel}.
During auto-regressive inference, the self-decoder maintains $S_n$ as the intermediate state for an efficient generation.

\mypara{The Chunkwise Recurrent Representation}
The chunk-wise representation is a unified formulation of recurrent and parallel representations.
Given chunk size $B$, the outputs are computed chunk by chunk.
The computation is divided into inner-chunk and cross-chunk parts.
Denote ${[i]}$ as the $i$-th chunk, i.e., $x_{[i]} = x_{(i-1)B+1} , \cdots , x_{iB}$, we compute the $i$-th chunk as:
\begin{equation}
\begin{aligned}
\label{eq:gret:chunk}
\quad \beta_{(i-1)B+j} &= \prod_{k=(i-1)B+1}^{(i-1)B+j} \gamma_{k},\quad D_{[i]}(j,k)=\frac{\beta_{(i-1)B+k}}{\beta_{(i-1)B+j}}\ \ \mathrm{if}\ \ j \le k\ \ \mathrm{else}\ \ 0 \\
R_{i} &= K_{[i]}^\intercal (V_{[i]}\odot \frac{\beta_{iB}}{\beta_{[i]}})+\beta_{iB} R_{i-1},\ \ \beta_{[i]}(j,k)= \beta_{(i-1)B+j}\\
\operatorname{gRet} (X) &= \underbrace{ (Q_{[i]} K^\intercal_{[i]} \odot D_{[i]}) V_{[i]} }_{\text{Inner-Chunk}} + \underbrace{ (Q_{[i]} R_{i-1}) \odot \beta_{[i]}}_{\text{Cross-Chunk}}
\end{aligned}
\end{equation}
where $R_i$ is the intermediate state of the $i$-th chunk, and $\beta$ summarizes the data-controlled decay $\gamma$.
The proof in \Cref{app:gret:chunk} shows the equivalence between the computation paradigms.
The chunkwise paradigm combines the best of parallelism and recurrence, i.e., saving FLOPs compared with fully parallel computation and reducing the iterations compared to recurrent computation.
During the training and prefill stages, the chunk-wise representation increases throughput and reduces GPU memory consumption.

\mypara{Multi-Head Gated Retention}
Similar to multi-head attention~\cite{transformer} and multi-scale retention~\cite{retnet}, we apply gated retention to each head and combine the outputs together:
\begin{equation}
\begin{aligned}
\mathrm{head}_i &= \operatorname{gRet} (X) \\
Y &= \operatorname{GroupNorm}_{h}( \operatorname{Concat}(\mathrm{head}_1, \cdots, \mathrm{head}_n) ) \\
\mathrm{MHGR}(X) &= (\operatorname{swish}(X W_G) \odot Y) W_O
\label{eq:mhgr}
\end{aligned}
\end{equation}
where $W_G, W_O \in \mathbb{R}^{d\times d}$ are learnable matrices, and $\operatorname{GroupNorm}$~\cite{groupnorm} normalizes each head~\cite{magneto}.
We also apply $\operatorname{swish}$ gate to increase non-linearity~\cite{retnet}.

\subsection{Sliding-Window Attention}
\label{sec:swa}

Sliding-window attention~\cite{sparsetransformer} restricts the attention range into a fixed window size $C$.
In contrast, vanilla Transformer decoders attend to all previous tokens.
During inference, the KV cache memory complexity can be reduced from $\mathcal{O}(N)$ to $\mathcal{O}(C)$, i.e., the memory usage is constant rather than increasing with sequence length.
Similar to multi-head self-attention~\cite{transformer}, we compute the output of sliding-window attention via:
\begin{equation}
\begin{aligned}
Q = X W_Q ,&\quad K = X W_K ,\quad V = X W_V \\
\mathrm{head}_i &= \softmax (Q_{[i]} K_{[i]}^\intercal + B) V \\
B_{ij} &=
\left\{
\begin{aligned}
& 0, &i - C < j \le i \\
& -\infty, &\text{otherwise} \\
\end{aligned}
\right. \\
Y &= \operatorname{Concat}(\mathrm{head}_1, \cdots, \mathrm{head}_h) \\
\mathrm{SWA}(X) &= Y W_O \\
\label{eq:swa}
\end{aligned}
\end{equation}
where $W_Q, W_K, W_V, W_O \in \mathbb{R}^{d\times d}$ are learnable matrices, and the window causal mask $B$ controls each query only attends to the previous keys whose distances are less than $C$.
The pre-normalization and residual connection are also applied to the module.

\section{Experiments}
\label{sec:exp}

We evaluate \our{} for large language models from the following perspectives.
First, we follow the setting of StableLM-3B-4E1T~\cite{stablelm} to scale up training tokens (\Cref{sec:lm:3b}).
Second, we present the scaling curves of the proposed architectures (\Cref{sec:scaling}).
Third, we scale up the \our{} model to 1M context length and evaluate its long-sequence modeling capability (\Cref{sec:long:eval}).
Fourth, we analyze the deployment advantages, including GPU memory footprint, serving capacity, prefilling time, and throughput (\Cref{sec:exp:infer}).
Experimental results show that \our{} achieves competitive performance across various evaluation metrics. More importantly, the proposed method significantly reduces the inference cost.

\subsection{Language Modeling Evaluation}
\label{sec:lm:3b}

We train a 3B-size \our{} language models by scaling up the number of training tokens. Then we compare the checkpoints with strong Transformer-based language models.

\mypara{Setup}
We use a similar training recipe as in StableLM-3B-4E1T~\cite{stablelm}.
We adjust the head dimension to 128 instead of 80 as in StableLM for better kernel support.
In order to keep the model size unchanged, we set the hidden size to 3072 and the number of layers to 26.
Grouped-query attention~\cite{gqa} is used, where the number of query heads is 24, and the number of key-value heads is 8.
We train \our{} with gated retention (\Cref{sec:gret}).
The non-embedding parameter count is 2.8B. In comparison, StableLM-3B-4E1T is 2.7B and OpenLLaMA-v2-3B~\cite{openllama} is 3.2B.
The training sequence length is 4096.
The batch size is 4M tokens.
We use the AdamW~\cite{adamw} optimizer with $\beta=0.9,0.95$.
The maximal learning rate is 3.2e-4 with 1000 warmup steps and linear decay to 1.28e-5.
The total schedule is set to 5T tokens. We train the model with 400k steps (i.e., 1.6T tokens) given the resource budget.
The curated training corpus is similar to \cite{stablelm}.
We use \texttt{tiktoken-cl100k\_base} as the tokenizer.
Detailed hyperparameters are described in \Cref{app:hp:3b}.

\begin{table*}[t]
\centering
\resizebox{\columnwidth}{!}{%
\begin{tabular}{@{}lccccccccc}
\toprule
\textbf{Model} & \textbf{ARC-C} & \textbf{ARC-E}& \textbf{BoolQ} & \textbf{Hellaswag} & \textbf{OBQA} & \textbf{PIQA} & \textbf{Winogrande} & \textbf{SciQ}  & \textbf{Avg} \\
\midrule
\multicolumn{10}{l}{\textit{Training with 1T tokens}} \\
OpenLLaMA-3B-v2 & 0.339 & 0.676 & \textbf{0.657} & \textbf{0.700} & 0.260 & 0.767 & 0.629 & \textbf{0.924} & 0.619 \\
StableLM-base-alpha-3B-v2 & 0.324 & 0.673 & 0.646 & 0.686 & 0.264 & 0.760 & 0.621 & 0.921 & 0.612 \\
StableLM-3B-4E1T  & --- & 0.666 & --- & --- & --- & \textbf{0.768} & 0.632 & 0.914 & ---  \\
\our{}-3B    & \textbf{0.379} & \textbf{0.731} & 0.645 & 0.689 & \textbf{0.298} & 0.763 & \textbf{0.639} & \textbf{0.924} &  \textbf{0.634} \\
\midrule
\multicolumn{10}{l}{\textit{Training with 1.6T tokens}} \\
StableLM-3B-4E1T  & --- & 0.688 & --- & --- & --- & 0.762 & 0.627 & 0.913 & ---  \\
\our{}-3B & 0.396 & 0.733 & {0.644} & 0.698 & 0.300 & 0.764 & 0.631 & 0.921 & 0.636 \\
\midrule
\multicolumn{10}{l}{\textit{Extending context length to 1M tokens}} \\
\our{}-3B-1M & {0.413} & {0.747} & 0.638 & {0.705} & 0.300 & {0.773} & {0.651} & {0.932} & {0.645} \\
\bottomrule
\end{tabular}
}
\caption{Eval Harness~\cite{eval-harness} results compared with previous well-trained Transformer language models~\cite{stablelm,stablelm-alphav2,openllama}.
We scale the 3B model to 1.6 trillion training tokens.
The 1T and 1.6T results of StableLM-3B-4E1T are taken from its technical report~\cite{stablelm}.
\our{}-3B-1M is extended to the context length of 1M tokens.
}
\label{tbl:harness}
\end{table*}

\mypara{Results}
\Cref{tbl:harness} compares the \our{} checkpoints with OpenLLaMA-v2-3B~\cite{openllama}, StableLM-base-alpha-3B-v2~\cite{stablelm-alphav2}, and StableLM-3B-4E1T~\cite{stablelm}.
We use LM Eval Harness~\cite{eval-harness} to evaluate the zero-shot performance on various downstream tasks.
OpenLLaMA-v2-3B and StableLM-base-alpha-3B-v2 are trained with 1T tokens.
The intermediate numbers of StableLM-3B-4E1T are taken from its technical report~\cite{stablelm}.
Experimental results across end tasks indicate that \our{} achieves comparable results with previous well-tuned Transformer language models.
Both the checkpoints trained with 1T tokens and 1.6T tokens obtain consistent trend.
Moreover, the results show that \our{} is scalable in terms of training tokens.

\subsection{Scalability Compared with Transformers}
\label{sec:scaling}

\begin{wrapfigure}{r}{0.524\textwidth}
\setlength\intextsep{0pt}
\centering
\vspace{-1.5em}
\captionsetup{type=figure}
\includegraphics[width=0.48\textwidth]{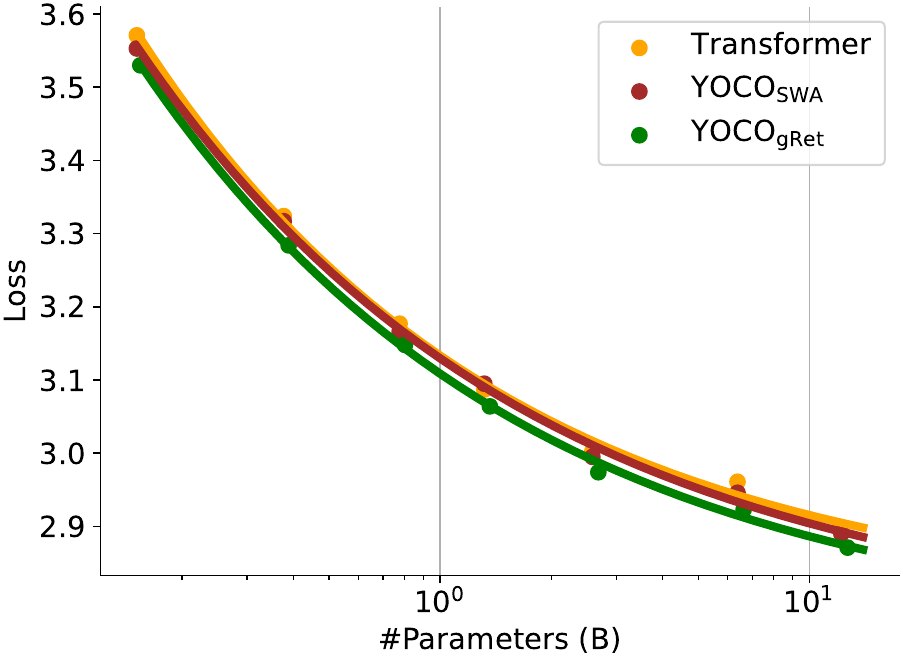}
\caption{LM loss decreases along with scaling up the model size (ranging from 160M to 13B).}
\vspace{-1.5em}
\label{fig:scaling}
\end{wrapfigure}

We compare the scaling curves between Llama Transformer~\cite{transformer,llama}, \our{} with gated retention (\our{}$_\text{gRet}$; \Cref{sec:gret}), and \our{} with sliding-window attention (\our{}$_\text{SWA}$; \Cref{sec:swa}).
We train language models of various sizes (i.e., 160M, 400M, 830M, 1.4B, 2.7B, 6.8B, and 13B) using the same training data and settings.
The validation loss is used as the evaluation metric.
The scaling law~\cite{scaling:law} is supposed to extrapolate larger-size performance.

\mypara{Setup}
We augment the Transformer architecture with Llama~\cite{llama} improvements, such as RMSNorm~\cite{rmsnorm}, SwiGLU~\cite{glu}, and removing bias.
The sliding window size of \our{}$_\text{SWA}$ is 1,024.
We align the number of parameters by adjusting the FFN intermediate dimension.
The training batch size is 0.25M tokens with a 2k sequence length.
We train the models with 40k steps, i.e., 10B tokens. In practice, we find that the setting is effective for loss convergence, and the scaling laws can be well-fitted.
More hyperparameters are detailed in \Cref{app:hp:scaling}.

\mypara{Results}
\Cref{fig:scaling} reports the validation loss with various parameter counts.
We also fit the scaling curves as in \cite{scaling:law}.
\our{} obtains comparable performance from 160M to 13B compared to the Llama-optimized transformer architecture.
The findings demonstrate that \our{} scales effectively with respect to model size.
Moreover, \our{}$_\text{gRet}$ outperforms Transformer and \our{}$_\text{SWA}$.
The gains come from hybrid architectures of attention and retention, whose inductive biases tend to be complementary to each other. We observed similar gains by interleaving the attention and retention modules (1:3). Recent hybrid architectures~\cite{jamba} also confirm similar findings.

\subsection{Long-Context Evaluation}
\label{sec:long:eval}

We extend the context length of \our{}-3B (\Cref{sec:lm:3b}) to 1M tokens.
We evaluate long-context models on needle retrieval and language modeling tasks.

We continue the model training with longer lengths progressively.
The length schedule is 64K, 256K, and 1M tokens.
The batch size is kept the same as before.
The learning rate and RoPE~\cite{rotary} $\theta$ are set as in \Cref{tbl:hp:length}.
Training data is up-sampled according to sequence length~\cite{length:upsampling}.
For a fair comparison, we do not use long-instruction tuning data. 
More training details are described in \Cref{app:hp:length}.
A chunk parallelism algorithm for \our{} is proposed in \Cref{app:chunk:parallelism}, which reduces communication overhead and GPU memory fragmentation in our experiments of 1M length.

\begin{figure*}[t]
\centering
\captionsetup{type=figure}
\includegraphics[width=0.8\textwidth]{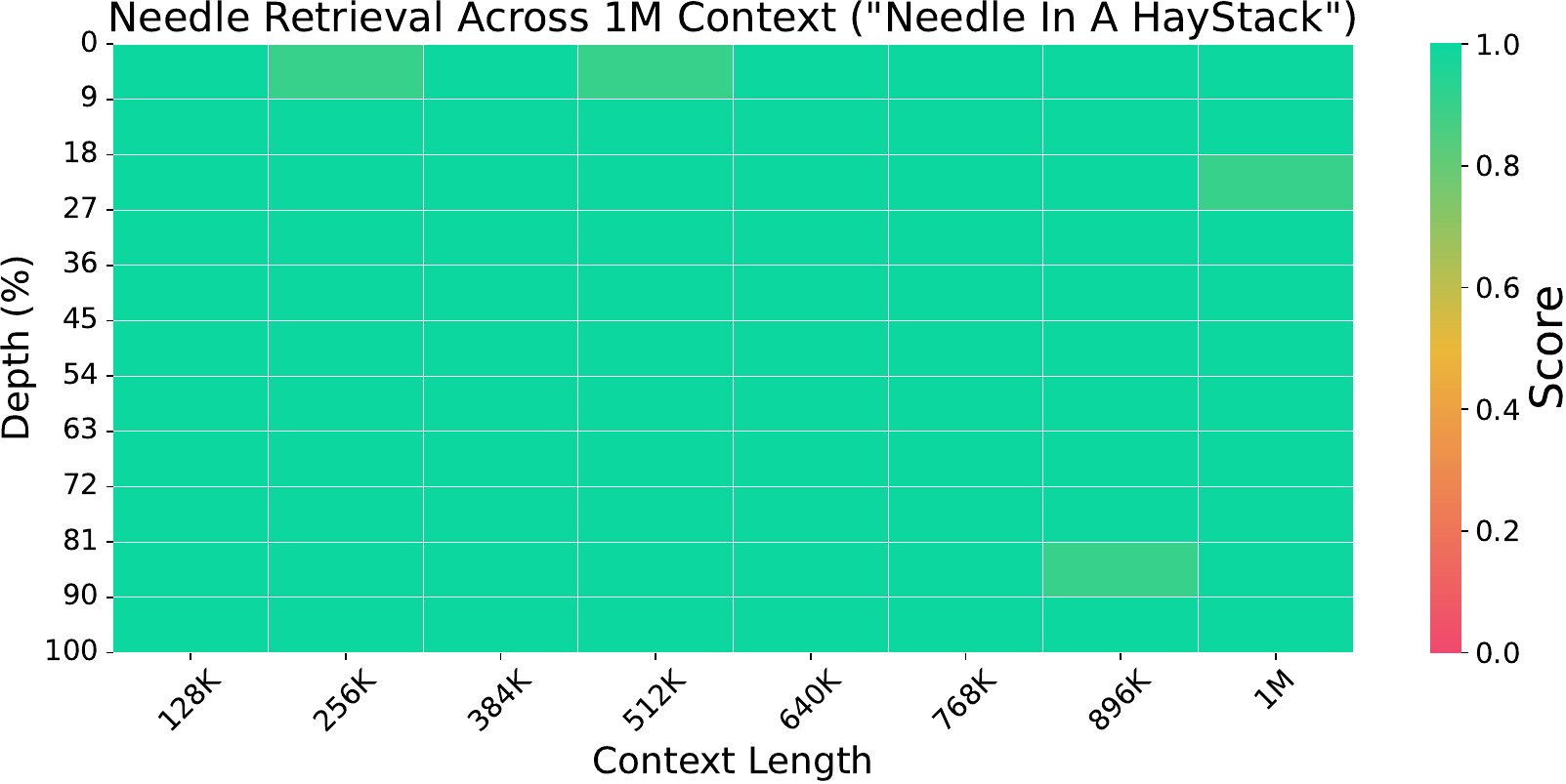}
\caption{Needle-in-a-haystack results in 1M length.
}
\label{fig:needle}
\end{figure*}

\begin{table*}[t]
\centering
\begin{tabular}{lccccc}
\toprule
\bf Model & \bf Size & $N=1$ & $N=2$ & $N=4$ & $N=8$ \\
\midrule
YaRN-Mistral-128K~\cite{yarn} & 7B & 0.02 & 0.12 & 0.08 & 0.20 \\
LWM-1M-text~\cite{lwm} & 7B & 1.00 & 0.90 & 0.76 & 0.62 \\
MiniCPM-128K~\cite{minicpm} & 2.4B & 1.00 & 1.00 & 0.54 & 0.56 \\
ChatGLM3-128K~\cite{glm} & 6B  & 0.94 & 0.72 & 0.52 & 0.44 \\
\our{}-3B-1M & 3B & 0.98 & 0.98 & 0.84 & 0.56 \\
\bottomrule
\end{tabular}
\caption{Multi-needle retrieval accuracy. $N$ indicates the number of needles. $N=1$ is single-needle retrieval used as a reference, and $N>1$ indicates the multi-needle test.
The evaluation is conducted in 128K length, because most previous long-context models are tuned with this length.
}
\label{tbl:multineedle}
\end{table*}

\mypara{Needle In A Haystack}
The pressure test evaluates whether models can retrieve ``needles'' from a long document~\cite{needle}.
We follow the evaluation setting of Gemini 1.5~\cite{gemini1.5} and LWM~\cite{lwm}.
The needles are constructed as a city with a magic number.
We run 10 times at the same depth and length. The averaged accuracy is reported.
\Cref{fig:needle} shows that \our{}-3B-1M passes the Needle-In-A-Haystack test with near perfect accuracy.
The results indicate that \our{} has strong long-context modeling capability.

\mypara{Multi-Needle Retrieval}
Besides the above single-needle retrieval, we conduct a multi-needle evaluation.
We compare \our{}-3B-1M with previous long-context language models, including MiniCPM-128K~\cite{minicpm}, ChatGLM3-128K~\cite{glm}, YaRN-Mistral-128K~\cite{yarn}, and LWM-1M-text~\cite{lwm}.
The evaluation is conducted in 128K sequence length, because most previous models are tuned with this length.

\Cref{tbl:multineedle} reports the accuracy with $N$ needles.
Among these models, LWM-1M-text and \our{}-3B-1M are trained with a 1M context length, while the others are in 128K length.
Although LWM-1M-text continues training of Llama-2-7B, \our{}-3B-1M can still achieve comparable performance with half the model size.
Moreover, the 7B-size YaRN-Mistral-128K~\cite{yarn} obtained by postion interpolation lags behind the other models.
Compared to MiniCPM-128K and ChatGLM3-128K, \our{}-3B-1M also outperforms these well-trained language models.

\begin{figure*}[t]
\centering
\begin{subfigure}[b]{0.495\textwidth}
\centering
\includegraphics[width=\textwidth]{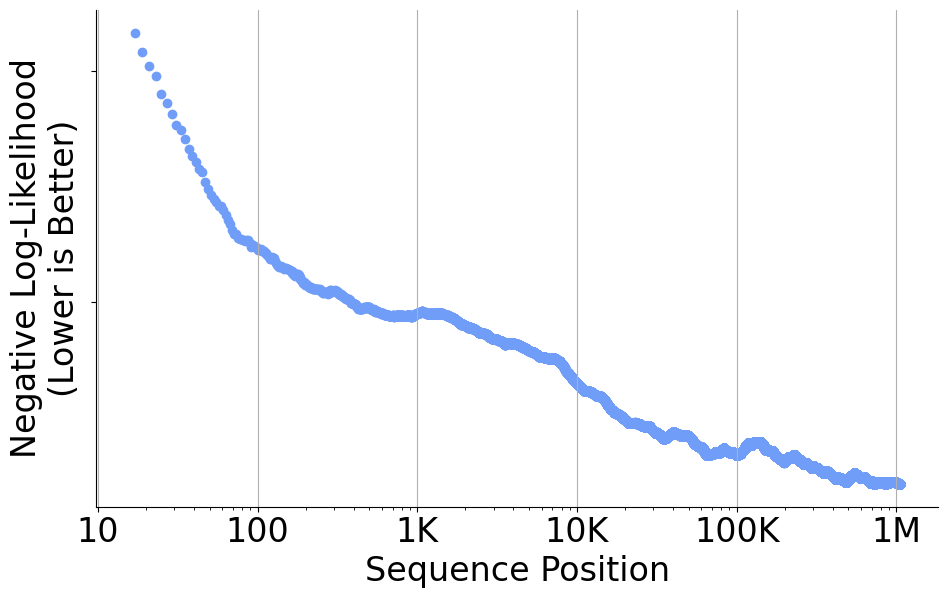}
\caption{Book data.}
\end{subfigure}
\hfill
\begin{subfigure}[b]{0.495\textwidth}
\centering
\includegraphics[width=\textwidth]{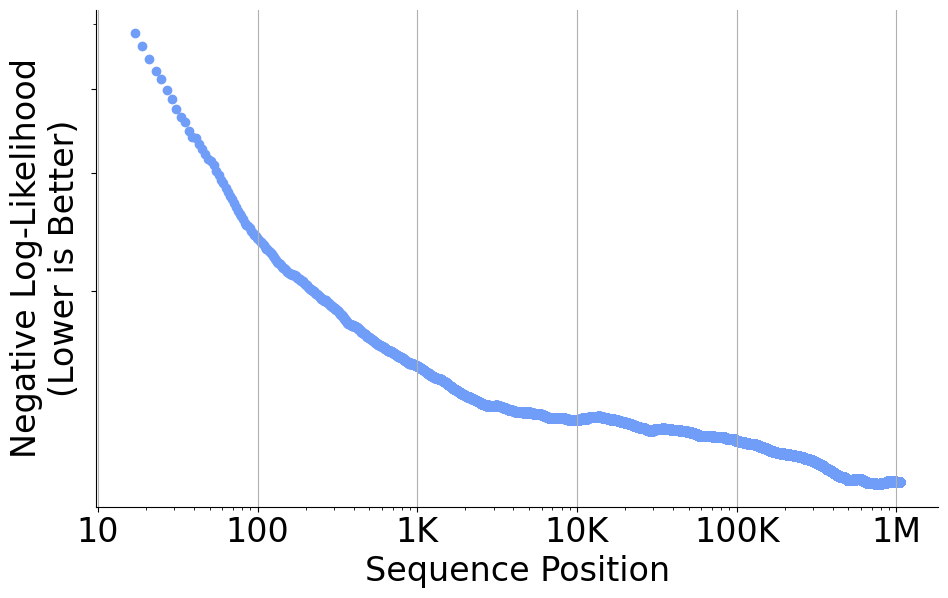}
\caption{Repository-level code data.}
\end{subfigure}
\caption{Cumulative average negative log-likelihood on book and repository-level code. We filter the validation examples that are longer than 1M tokens. \our{} achieves improved performance with longer context, i.e., utilizing long-distance information for language modeling.}
\label{fig:longppl}
\end{figure*}

\mypara{Perplexity over Long Sequences}
\Cref{fig:longppl} shows the cumulative average negative log-likelihood (NLL) as a function of context length.
We evaluate both book and repository-level code data.
We follow the setting of \cite{gemini1.5} and filter validation data that are longer than 1M tokens.
NLL decreases consistently with longer sequence length.
The results indicate that \our{} can effectively utilize long-distance dependency for language modeling.
We also observe that the NLL-length curves tend to fit the power law, where the gaps are affected by the noise within the validation examples.

\subsection{Inference Advantages}
\label{sec:exp:infer}

We analyze inference efficiency from various perspectives, such as GPU memory footprint, prefilling latency, throughput, and serving capacity.
We demonstrate that \our{} reduces the deployment cost by orders of magnitude, especially for long-sequence inference.
More importantly, the user experience (such as latency) is improved while maintaining good performance and reducing expenses.

We compare \our{}$_\text{gRet}$ with Transformer. The default model configuration follows \Cref{sec:lm:3b}.
Notice that Transformer uses grouped-query attention~\cite{gqa}, Flash-Decoding~\cite{flashdec}, and kernel fusion for a fair comparison.
As described in \Cref{sec:gret}, gated retention uses the chunk-recurrent representation in the prefill stage, and the recurrent representation in the generation stage.
The chunk size is set to 256.
We implement a Triton~\cite{triton} kernel for gated retention.
The evaluation sequence length is ranging from 32K to 1M.
The last 1,024 tokens are supposed to be generated, while the previous tokens are given input context.
The experiments are conducted with H100-80GB GPU cards.

\begin{figure*}[t]
\centering
\begin{subfigure}[c]{0.7\textwidth}
\centering
\includegraphics[width=\textwidth]{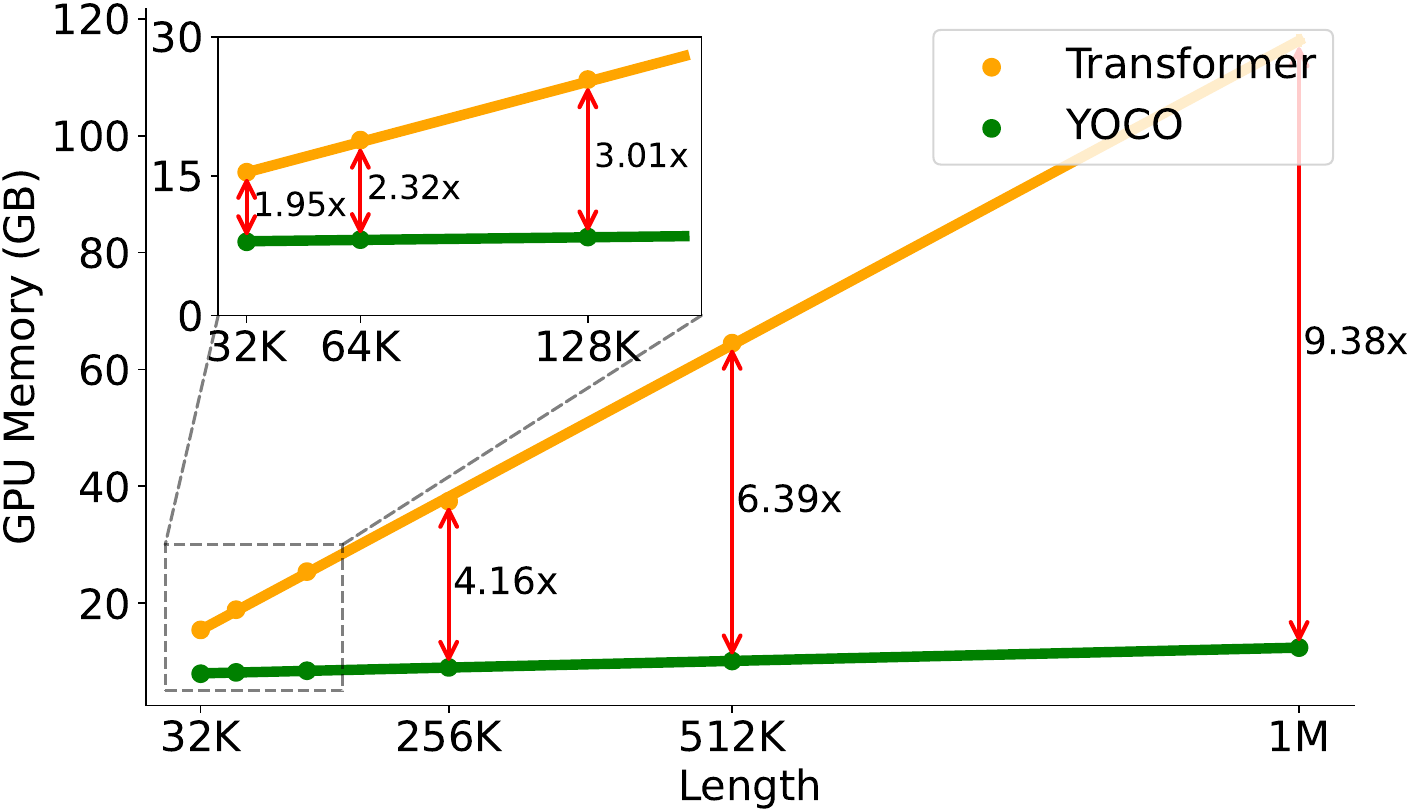}
\caption{Inference memory of Transformer and \our{} across various lengths.
}
\label{fig:memory-length}
\end{subfigure}
\hfill
\begin{subfigure}[c]{0.28\textwidth}
\centering
\includegraphics[width=\textwidth]{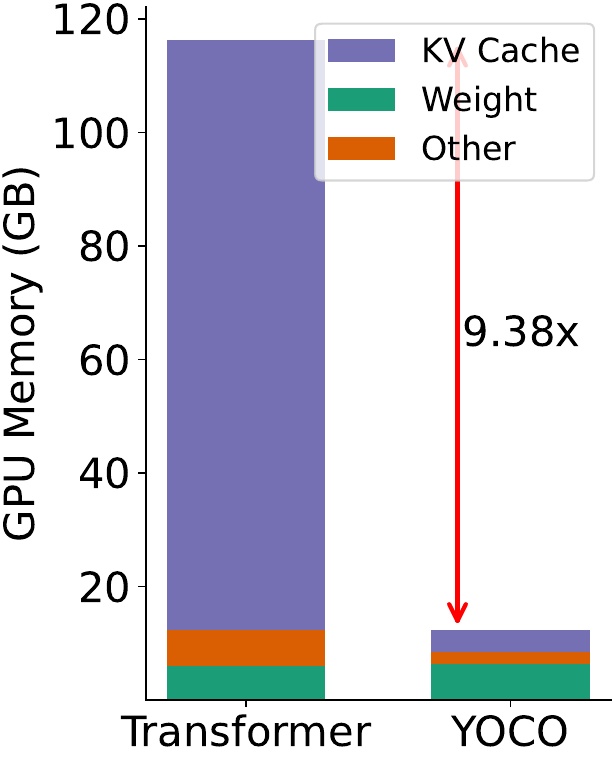}
\caption{Breakdown memory consumption in 1M context length.
}
\label{fig:memorybar}
\end{subfigure}
\caption{GPU memory consumption during inference.}
\label{fig:memory}
\end{figure*}

\mypara{GPU Memory}
The inference memory consumption is made up of three parts, namely model weights, intermediate activation, and KV cache.
\Cref{fig:memorybar} presents the breakdown memory profiling results.
Along with an increase in context length, the main memory bottleneck becomes KV caches, while model weights consume constant memory.
The results show that \our{}$_\text{gRet}$ alleviates the activation cost and KV cache memory footprint.

As shown in \Cref{fig:memory-length}, the memory cost is significantly reduced using \our{}.
Moreover, the memory consumption of \our{} increases slowly along the sequence length.
For example of 1M length, the overall inference memory usage is only 12.4GB, while Transformers occupy $9.4\times$ GPU memory.
\our{} makes it feasible to deploy long-sequence modeling on customer-level GPUs.
Even with a 32K sequence length, \our{} requires about $2\times$ less memory than Transformer.
Although we compare 3B-size models here, the reduction ratio becomes larger as the number of layers increases.

\begin{figure*}[t]
\centering
\captionsetup{type=figure}
\includegraphics[width=0.64\textwidth]{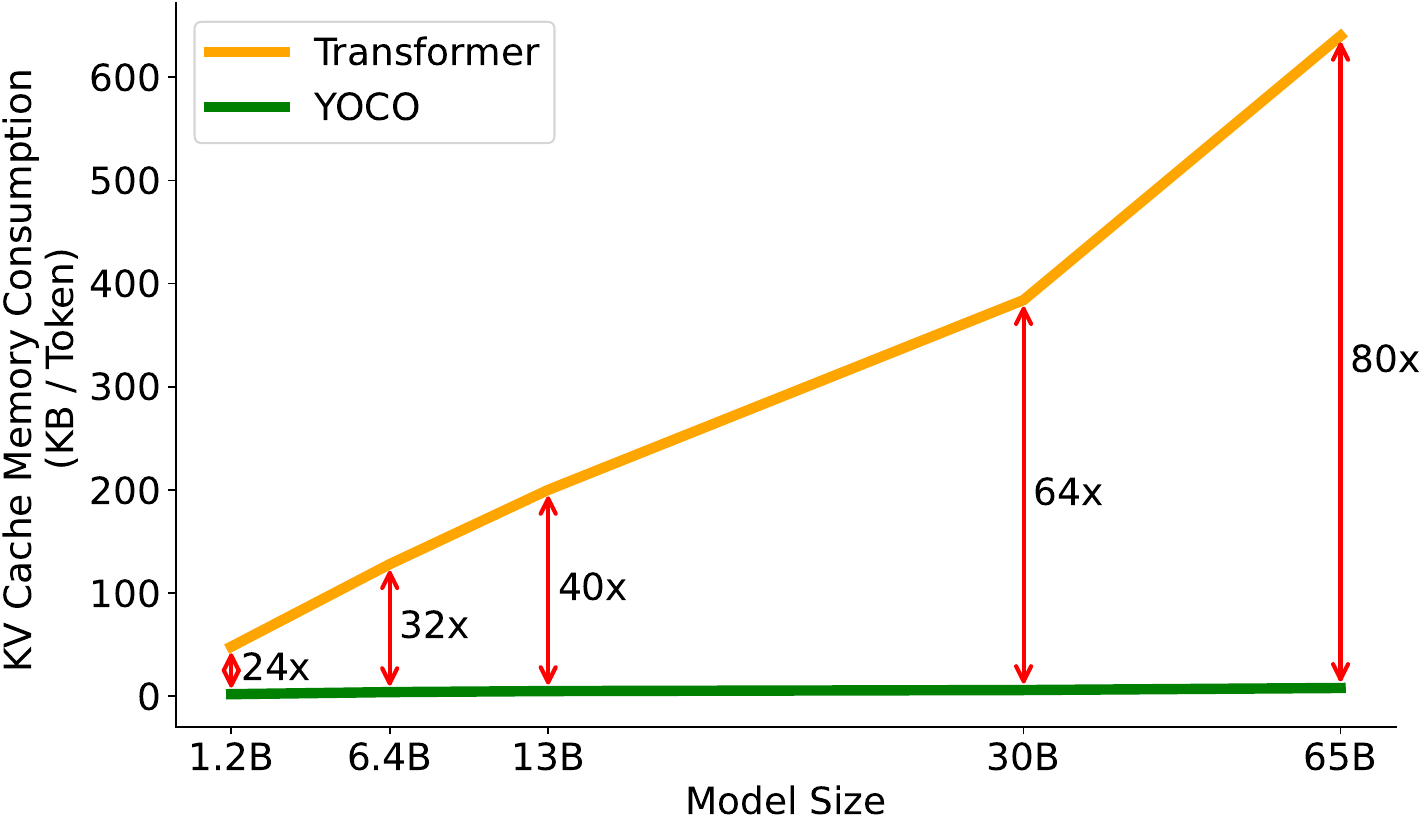}
\caption{GPU memory consumption of KV cache for each token with different model size. \our{} can save more for larger model size.}
\label{fig:cache:mem:per:token}
\end{figure*}

\Cref{fig:cache:mem:per:token} reports the GPU memory consumption of KV cache for each token.
As \our{} only caches one layer of global key-value pairs, it needs roughly $L$ times fewer memory compared to Transformer.
For example, \our{} can serve 128K tokens with 1GB GPU memory, while Transformer with GQA~\cite{gqa} can only support 1.6K tokens at 65B model size.

\begin{figure*}[t]
\centering
\captionsetup{type=figure}
\includegraphics[width=0.75\textwidth]{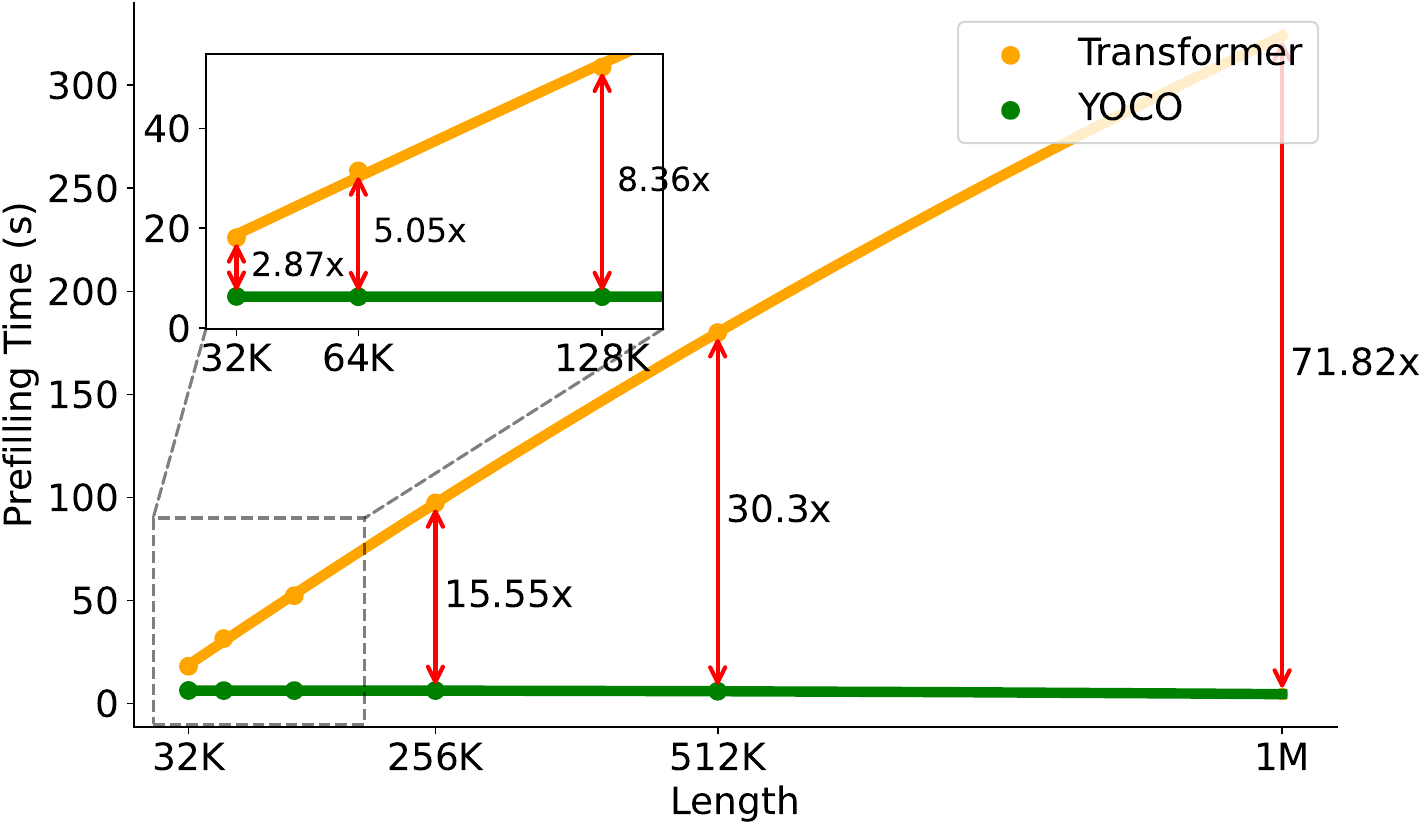}
\caption{Prefilling latency for different length, i.e., the encoding time of given input prompt before generating the first token. Transformer's time grows quadratically while \our{}'s grows linearly.
Even for a short input length, such as 32K, \our{} can still accelerate $2.87\times$.
}
\label{fig:prefilling}
\end{figure*}

\mypara{Prefilling Latency}
In the prefill stage, the model encodes input tokens in parallel.
As shown in \Cref{fig:prefilling}, the prefilling latency is a pain point of user experience for long-context models.
For 512K- and 1M-length input sequences, Transformer needs about 180 seconds and 300 seconds, respectively.
The computational complexity of Transformer is $\mathcal{O}(N^2)$, which requires a large number of FLOPs for long context.
In contrast, \our{}'s prefilling time is $\mathcal{O}(N)$, growing linearly (\Cref{sec:infer:adv}) along the sequence length.

\Cref{fig:prefilling} shows that \our{} reduces the Transformer prefilling time from 180 seconds to less than 6 seconds for 512K context.
As described in \Cref{sec:infer:adv}, the prefill stage can early exit before entering cross-decoder. So, there is at least two times speedup of prefilling latency even for short context.
For example, \our{} is $2.87\times$ faster than Transformer for 32K length.

\begin{figure*}[t]
\centering
\captionsetup{type=figure}
\includegraphics[width=0.6\textwidth]{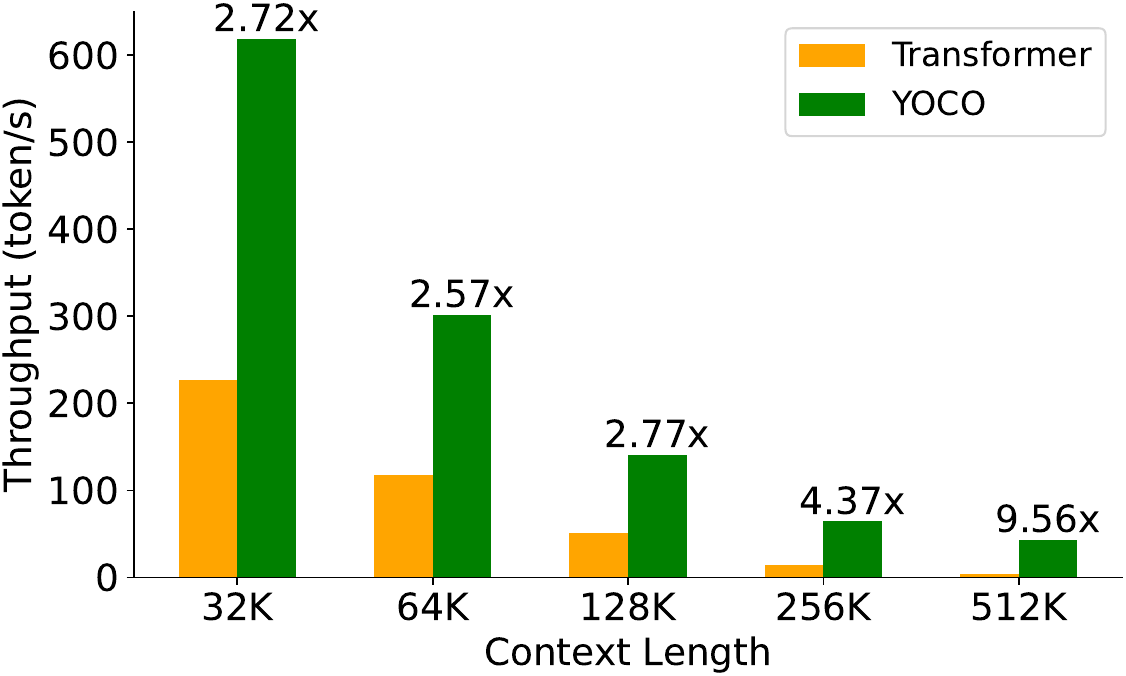}
\caption{Inference throughput of Transformer and \our{} varying the context length.}
\label{fig:throughput}
\end{figure*}

\mypara{Throughput}
The throughput indicates how many tokens the model can process per second, involving both pre-filling and generation time.
\Cref{fig:throughput} shows that \our{} achieves higher throughput across context lengths compared to Transformer.
For the example of 512K queries, Transformer's throughput is 4.5 token/s while \our{} reaches 43.1 token/s, i.e, achieving $9.6\times$ speedup.
The throughput is improved for the following reasons.
First, \our{} decreases the time required for prefilling as previously demonstrated.
Second, as the memory consumption is reduced, we can use larger batch size for inference, which also contributes to the throughput improvement.

\section{Conclusion}

In this work, we propose a decoder-decoder architecture (\our{}) for large language modeling.
\our{} achieves significantly better inference efficiency and competitive performance compared with Transformers.
Experimental results demonstrate that \our{} achieves favorable results for large language models under various settings, i.e., scaling up number of training tokens, scaling up model size, and scaling up context length to 1M tokens.
Profiling results also show that \our{} improves inference efficiency by orders of magnitude, especially for long-sequence modeling.

The work can be advanced from the following perspectives:
\begin{itemize}[leftmargin=*]
\setlength\itemsep{0.01em}
\item \textbf{\our{} + BitNet + Groq.} Groq achieves very high throughput by putting all things within SRAM. However, the memory capacity bottleneck limits the model size and input token count. Now, hundreds of chips are connected to host just one model. As a solution, \our{} reduces KV cache memory, and BitNet reduces model weight memory. The LLM deployment cost is expected to be reduced by orders of magnitude using the above combination.
\item \textbf{\our{} for Multimodal Large Language Models.} The \our{} layout is general to the use of multiple self-decoders. The cross-attention layers are natural for multimodal fusion~\cite{vlmo,beit3}. The causal dependency of self-decoders also perfectly fits in streaming video. The async multimodal large language models can avoid different data steams block each other, which is critical for real-time applications, such as robotics.
\item \textbf{Optimized Mechanism for KV Cache Module.} \Cref{fig:yoco} explicitly highlights KV cache, which opens up new opportunities to develop native memory mechanisms.
First, we can integrate a cache compression mechanism to obtain more compact memory. Second, we can build an index~\cite{longmem} for efficient key-value retrieval. As \our{} reuses caches, it enables us to maintain only one index rather than creating an index for each layer. Third, the disentangled modeling supports pre-caching context, which is potentially useful for native RAG and LLM-native search engines.
\end{itemize}

\section*{Acknowledgement}

We would like to acknowledge Ben Huntley for maintaining the GPU cluster.
The long-sequence training utilizes \texttt{CUBE}, which is an internal version of \cite{cube}.
We implement the Triton kernel of gated retention based on \texttt{FLA}~\cite{fla}.

\bibliographystyle{alpha}
\bibliography{arch}

\newpage
\appendix

\section{Chunk Parallelism for Long-Sequence Training of YOCO}
\label{app:chunk:parallelism}

We introduce chunk parallelism for \our{} to reduce the communication frequency, accelerating long-sequence training.
Dividing long sequences into different devices is essential when the training length is extremely long~\cite{seqparallel, longnet}.
However, the overall throughput tends to be bounded by GPU communication~\cite{ringattn}.
Cross-decoder disentangles self-attention dependency while preserving modeling capability, bringing intriguing advantages to distributed long-sequence training.

\begin{figure*}[ht]
\centering
\captionsetup{type=figure}
\includegraphics[width=0.6\textwidth]{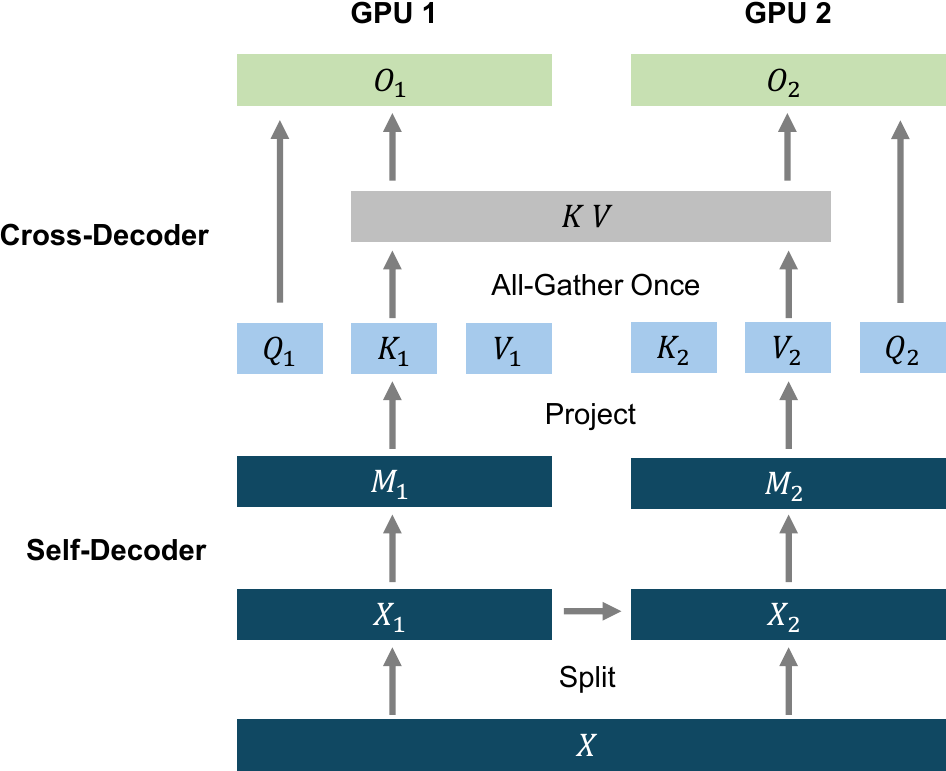}
\caption{Chunk parallelism of \our{} training on two GPU devices. The training strategy is to partition the sequence into different chunks. $M$ denotes the intermediate representation $X^{\nicefrac{L}{2}}$, i.e., the output of self-decoder.
The keys and values in the cross-decoder are only gathered once.}
\label{fig:chunk:parallel}
\end{figure*}

In self-decoder, the dependency only exists in the adjacent devices.
For example, gated retention only requires the hidden state $S_n$ in \Cref{eq:gret:recurrent}, and sliding-window attention attends to tokens within the context window.
Therefore, the communication amount of self-decoder is relatively small.
In the cross-decoder, the all-gather operation is only triggered once for the KV cache, rather than communicating in each layer.
The hardware-friendly architecture gives more flexibility to distributed long-sequence training.


\section{Chunk-wise Representation of Gated Retention}
\label{app:gret:chunk}

We illustrate the equivalence between recurrent representation and chunkwise recurrent representation of gated retention.
For the output $O_n$, $n$ can be split as $n=kB + r$ where $B$ is the chunk size:
\begin{equation}
\begin{aligned}
\label{eq:gret:chunkproof1}
O_n =\sum_{m=1}^n \prod_{i=m+1}^{n}\gamma_i Q_nK_m^\intercal V_m &= \sum_{m=kB+1}^n \prod_{i=m+1}^{n}\gamma_i Q_nK_m^\intercal V_m+\sum_{m=1}^{kB}\prod_{i=m+1}^{n}\gamma_i Q_nK_m^\intercal V_m \\
\sum_{m=kB+1}^n \prod_{i=m+1}^{n}\gamma_i Q_nK_m^\intercal V_m &= (Q_n K_{kB+1:n}^\intercal\odot \Gamma_{kB+1:n}) V_{kB+1:n} \\
\sum_{m=1}^{kB}\prod_{i=m+1}^{n}\gamma_i Q_nK_m^\intercal V_m &= (Q_n\prod_{i=kB+1}^{n}\gamma_i) \sum_{c=0}^{k-1} \sum_{m=1}^{B} (K_{m+cB}^\intercal V_{m+cB} \prod_{i=m+cB+1}^{(c+1)B}\gamma_i) \prod_{i=(c+1)B+1}^{kB}\gamma_i\\
 & = (Q_n\prod_{i=kB+1}^{n-1}\gamma_i) \sum_{c=1}^{k} (K_{[c]}^\intercal (V_{[c]}\odot \zeta_{[c]})) \prod_{i=c+1}^{k}\alpha_i \\
 & = (Q_n\prod_{i=kB+1}^{n-1}\gamma_i) R_{i-1} \\
\end{aligned}
\end{equation}
where $\Gamma_i=\prod_{k=i+1}^{n}\gamma_i$, $\zeta_{[c]}(j,k)= \prod_{i=(c-1)B+j+1}^{cB}\gamma_i$, $\alpha_{i}= \prod_{j=(i-1)B+1}^{iB}\gamma_j$, ${[i]}$ indicates the $i$-th chunk, i.e., $x_{[i]} = [x_{(i-1)B+1} , \cdots , x_{iB}]$. $R_n$ is written as a recurrent function:
\begin{equation}
\label{eq:gret:chunkproof2}
\begin{aligned}
    R_i=K_{[i]}^\intercal (V_{[i]}\odot \zeta_{[i]}) + \alpha_i R_{i-1}
\end{aligned}
\end{equation}
Denote ${[i]}$ as the $i$-th chunk, i.e., $x_{[i]} = [x_{(i-1)B+1} , \cdots , x_{iB}]$, $\beta_{(i-1)B+j} = \prod_{k=(i-1)B+1}^{(i-1)B+j}$, $\beta_{[i]}(j,k)= \beta_{(i-1)B+j}$, We concatenate the output in a block together:
\begin{equation}
\begin{aligned}
\label{eq:gret:chunkproof3}
O_{[n]} &= \sum_{m=kB+1}^{[n]} \beta_{[n]} Q_{[n]}K_m^\intercal V_m + \sum_{m=1}^{kB} \beta_{[n]} Q_{[n]} \prod_{i=m+1}^{n}\gamma_i K_m^\intercal V_m \\
\sum_{m=kB+1}^{[n]} \beta_{[n]} Q_{[n]}K_m^\intercal V_m&=(Q_{[n]} K^\intercal_{[n]}\odot D_{[n]}) V_{[n]} ,\quad D_{[n]}(j,k)=\frac{\beta_{(n-1)B+k}}{\beta_{(n-1)B+j}}\ \ \mathrm{if}\ \ j \le k\ \ \mathrm{else}\ \ 0 \\
\sum_{m=1}^{kB} \beta_{[n]} Q_{[n]} \prod_{i=m+1}^{n}\gamma_i K_m^\intercal V_m &= \beta_{[n]} Q_{[n]} R_{i-1}, \quad R_{i}=K_{[i]}^\intercal (V_{[i]}\odot \frac{\beta_{iB}}{\beta_{[i]}})+\beta_{iB} R_{i-1}, \\
O_{[n]} &= \underbrace{ (Q_{[n]} K^\intercal_{[n]}\odot D_{[n]}) V_{[n]} }_{\text{Inner-Chunk}} + \underbrace{ (Q_{[n]}R_{n-1}) \odot \beta_{[n]}}_{\text{Cross-Chunk}}
\end{aligned}
\end{equation}

Finally, we show that the chunkwise recurrent representation of gated retention is equivalent to the other two representations.

\section{Hyperparameters for \our{}-3B}
\label{app:hp:3b}

We describe the hyperparameters used for \Cref{sec:lm:3b}.
The hidden dimension is set to 3072.
The number of layers is 26.
The number of query heads is 24, and the number of key/value heads is 8 with grouped-query attention~\cite{gqa}.
The total number of parameters without embedding is 2.83B.
The training batch size is 4M tokens.
We use 4096 training length.
The optimizer is AdamW~\cite{adamw} with $\beta = (0.9, 0.95)$. The learning rate is $3.2\times10^{-4}$ with 1000 warmup steps.
We set a 5T-token learning rate schedule with linear decay to $1.28\times10^{-5}$.

\begin{table}[ht]
\centering
\begin{tabular}{lc}
\toprule
\textbf{Params} & \textbf{Values} \\
\midrule
Layers  & {26} \\
Hidden size & {3072} \\
FFN size & {8192} \\
Vocab size & 100,288 \\
Heads & {24} \\
Key-value heads & {8} \\
Adam $\beta$ & {(0.9, 0.95)} \\
LR & $3.2\times10^{-4}$ \\
Batch size & {4M} \\
Warmup steps & {1000} \\
Weight decay & {0.1} \\
Dropout & {0.0} \\
\bottomrule
\\
\end{tabular}
\caption{Hyperparamters used for the \our{}-3B model in~\Cref{sec:lm:3b}.
}
\label{tbl:hp:3b}
\end{table}

\section{Hyperparameters for Scaling Curves}
\label{app:hp:scaling}

We describe the hyperparameters used for \Cref{sec:scaling}.
\Cref{tbl:hp:scaling} reports the hidden dimension, number of layers, and number of heads used for different model sizes.
The head dimension of gated retention is set to 256.
To align the number of parameters, the FFN size for Transformer is $\frac{8}{3}d$ while the FFN size for \our{} is $3d$.
The training length is set to 2048.
The batch size is set to 0.25M tokens.
We use the AdamW~\cite{adamw} optimizer with $\beta_1=0.9,\beta_2=0.98$.
The learning rate is $1.5\times10^{-4}$ for 160M to 1.4B sizes and $7.5\times10^{-5}$ for 2.7B to 13B sizes.
The warmup step is 375 with linear rate decay. The weight decay is set to 0.05.
We train the models with 40k steps, i.e., 10B tokens.

\begin{table}[ht]
\centering
\begin{tabular}{lccc}
\toprule
\bf Size & \bf Hidden Dim. & \bf \#Layers & \bf \#Heads \\
\midrule
160M & 768 & 12 & 12 \\
400M & 1024 & 24 & 16\\
830M & 1536 & 24 & 12 \\
1.4B & 2048 & 24 & 16 \\
2.7B & 2560 & 32 &  20 \\
6.8B & 4096 & 32 & 32 \\
13B & 5120 & 40 & 40 \\
\bottomrule
\\
\end{tabular}
\caption{Model size and hyper-parameters used for scaling curves in~\Cref{sec:scaling}.}
\label{tbl:hp:scaling}
\end{table}

\section{Hyperparameters for Length Extension}
\label{app:hp:length}

We progressively extend the context length to 1M tokens in \Cref{sec:long:eval}.
The length schedule is 64K, 256K, and 1M.
We up-sample the documents that are longer than the training length.
\Cref{tbl:hp:length} shows that we use different RoPE $\theta$ and learning rate for each stage.

\begin{table}[ht]
\centering
\begin{tabular}{lccc}
\toprule
Training Length  & 65,536 & 262,144 & 1,048,576 \\
\midrule
Learning Rate  & $8\times10^{-5}$ & $4\times10^{-5}$ & $2\times10^{-5}$ \\
RoPE $\theta$  & 640K & 5M & 80M \\
Training Tokens & 6B & 4B & 1.5B \\
\bottomrule
\\
\end{tabular}
\caption{Hyperparamters used for length extension in~\Cref{sec:long:eval}.
}
\label{tbl:hp:length}
\end{table}

\section{Pseudo Code of Gated Retention}

We present pseudocode for the three computation paradigms of gated retention (\Cref{sec:gret}).
Parallel implementation enables training parallelism to fully utilize GPUs. The recurrent paradigm enables low-cost inference. Chunkwise retention combines the above advantages (i.e., parallel within each chunk and recurrent across chunks), which has linear memory complexity for long sequences.

\begin{minipage}[t]{\linewidth}
\centering
\begin{lstlisting}[language=python, mathescape, breaklines=true]  
def ParallelRetention(
    q, # bsz * num_head * len * dim
    k, # bsz * num_head * len * dim
    v, # bsz * num_head * len * dim
    gt): # bsz * num_head * len
    retention = q @ k.transpose(-1, -2)
    causal_mask = torch.full([q.shape[-2], q.shape[-2]], float("-inf"), device=q.device).triu(1).type_as(q)
    gt = F.logsigmoid(gt).cumsum(-1) / gate_logit_normalizer
    mask = (g[..., None] - g[..., None, :] + causal_mask).exp()
    
    retention = retention * mask
    output = retention @ v
    output = group_norm(output)
    return output
\end{lstlisting}
\end{minipage}

\begin{minipage}[t]{\linewidth}
\centering
\begin{lstlisting}[language=python, mathescape, breaklines=true]
def RecurrentRetention(
    q, k, v, # bsz * num_head * dim
    past_kv, # bsz * num_head * dim * dim
    gt # bsz * num_head * 1 * 1
    ):
    gt = F.logsigmoid(gt) / gate_logit_normalizer
    current_kv = gt.exp() * past_kv + k.unsqueeze(-1) * v.unsqueeze(-2)
    output = torch.sum(q.unsqueeze(-1) * current_kv, dim=-2)
    output = group_norm(output)
    return output, current_kv
\end{lstlisting}
\end{minipage}

\begin{minipage}[t]{\linewidth}
\centering
\begin{lstlisting}[language=python, mathescape, breaklines=true]
def ChunkwiseRetention(
    q, k, v, # bsz * num_head * chunk_size * dim
    past_kv, # bsz * num_head * dim * dim
    gt): # bsz * num_head * chunk_size
    gt = F.logsigmoid(gt).cumsum(-1) / gate_logit_normalizer
    cross_retention = (q @ past_kv) * gt[..., None].exp()
    inner_retention = ParallelRetention(q, k, v, gt)
    retention = inner_retention + cross_retention
    output = group_norm(retention)
    
    value_decay = (-gt + gt[:, :, :, -1, None]).exp()[..., None]
    chunk_decay = gt[..., -1].exp()
    current_kv = chunk_decay * past_kv + k.transpose(-1, -2) @ (v * value_decay)
    return output, current_kv
\end{lstlisting}
\end{minipage}

\section{Comparisons with Transformer Variants}
\label{sec:exp:variants}

We compare \our{}$_\text{gRet}$ and \our{}$_\text{SWA}$ with Transformer and other variants, including H3~\cite{h3}, RetNet~\cite{retnet}, Mamba~\cite{mamba}, and \gretnet{} (\Cref{sec:gret}).
All models have 160M parameters with 12 layers and a hidden dimension of 768.
The weights of word embedding and $\softmax$ projection are shared.
For Mamba, we follow all the details in the paper~\cite{mamba}, where double-SSM layers are implemented instead of ``SSM + SwiGLU''.
For H3, the experiment uses a hybrid version following the original paper~\cite{h3}, where attention layers are inserted into the second layer and the $\frac{L}{2}+1$ layer.
For RetNet and \gretnet{}, the value dimension is $d$ instead of $2d$, and the intermediate dimension of SwiGLU is $\frac{7}{3}d$ to match the number of parameters.

\subsection{Fine-Grained LM Perplexity Results}

\Cref{tbl:ablation:ppl} reports the validation perplexity for language modeling.
Following Zoology~\cite{zoology}, we divide the perplexity into \textbf{Ar-Hit}, where the predicted token is a bigram previously seen in the previous context, and \textbf{First-Occur}, where the predicted token cannot be recalled from the context.

\begin{table*}[ht]
\centering
\begin{tabular}{@{}lccc}
\toprule
   & \textbf{Valid. Set}      &  \textbf{AR-Hit}   &   \textbf{First-Occur} \\
\midrule
Mamba~\cite{mamba} & 3.645 & 1.555 & 4.126 \\
RetNet~\cite{retnet} & 3.633 & 1.466 & 4.131  \\
Hybrid H3~\cite{h3} & 3.591 & 1.251 & 4.130 \\
\gretnet{} & 3.600 & 1.354 & 4.116 \\
\midrule
Transformer & 3.564 & 1.219 & 4.104 \\
YOCO$_{\mathrm{SWA}}$ & 3.553 & 1.202 & 4.094 \\
YOCO$_{\operatorname{gRet}}$ & \textbf{3.530} & \textbf{1.199} & \textbf{4.067} \\
\bottomrule
\end{tabular}
\caption{Fine-grained perplexity results on language modeling. We report perplexity on both the overall validation set and the fine-grained diagnosis sets~\cite{zoology}, i.e., ``AR-Hit'' evaluates the associative recall capability, and ``First-Occur'' indicates the regular language modeling performance.
}
\label{tbl:ablation:ppl}
\end{table*}

\subsection{Long-Context Evaluation}

We evaluate the long-context modeling for the above architectures on four tasks of the ZeroSCROLLS~\cite{zeroscrolls} benchmark.
We continue training the 160M models in \Cref{tbl:ablation:ppl} as long-context models.
Specifically, we further train the models with 2B tokens in 16,384 length.
The rotation base scaling~\cite{long-meta} is also used for length extension.
For sparse Transformer, we keep the 2,048 context window and do not change the rotation base (i.e., RoPE $\theta$).

\begin{figure*}[ht]
\centering
\includegraphics[width=0.72\textwidth]{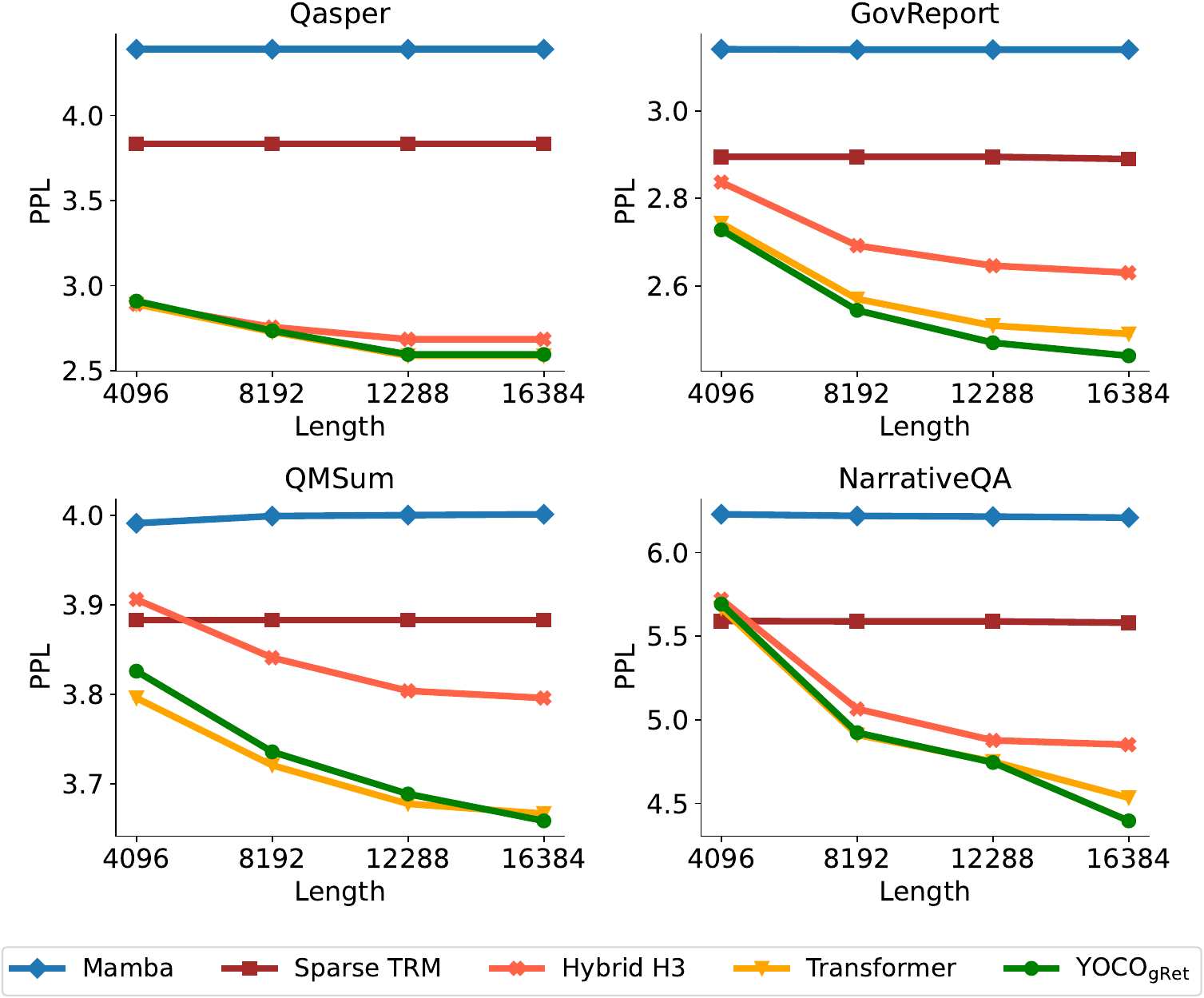}
\caption{Long sequence task perplexity decreases along with the increasing input length.}
\label{fig:scrolls}
\end{figure*}

\Cref{fig:scrolls} reports the perplexity of the answers with different input lengths.
Among all these architectures, \our{} and Transformer consistently perform better than others across tasks and lengths.

\end{document}